\documentclass{bmvc2k}

\usepackage{graphicx}
\usepackage{amsmath}
\usepackage{amssymb}
\usepackage{booktabs}
\usepackage{wrapfig}
\usepackage{blindtext}
\usepackage{multirow}
\usepackage{arydshln}
\usepackage{makecell}
\usepackage[normalem]{ulem}
\usepackage{bbold}
\usepackage{tikz}
\usepackage{comment}
\usepackage{amsmath,amssymb} 
\usepackage{color}

\usepackage[accsupp]{axessibility}  

\newcommand{\etal}{\textit{et al}.~}
\newcommand{\ie}{\textit{i}.\textit{e}.~}
\newcommand{\eg}{\textit{e}.\textit{g}.~}

\usepackage[capitalize]{cleveref}
\crefname{section}{Sec.}{Secs.}
\Crefname{section}{Section}{Sections}
\Crefname{table}{Table}{Tables}
\crefname{table}{Tab.}{Tabs.}

\title{Detailed Annotations of Chest X-Rays via CT Projection for Report Understanding
}

\addauthor{Constantin Seibold}{constantin.seibold@kit.edu}{1}
\addauthor{Simon Rei\ss}{simon.reiss@kit.edu}{1}
\addauthor{Saquib Sarfraz}{muhammad.sarfraz@kit.edu}{1,2}
\addauthor{Matthias A. Fink}{matthias.fink@uni-heidelberg.de}{3}
\addauthor{Victoria Mayer}{victoria.mayer@med.uni-heidelberg.de}{3}
\addauthor{Jan Sellner}{j.sellner@dkfz-heidelberg.de}{4}
\addauthor{Moon Sung Kim}{moon-sung.kim@uk-essen.de}{5}
\addauthor{Klaus H. Maier-Hein}{k.maier-hein@dkfz-heidelberg.de}{4}
\addauthor{Jens Kleesiek}{jens.kleesiek@uk-essen.de}{4}
\addauthor{Rainer Stiefelhagen}{rainer.stiefelhagen@kit.edu}{1}

\addinstitution{
 Institute of Anthropomatics and Robotics\\
 Karlsruhe Institute of Technology\\
 Karlsruhe, Germany
}
\addinstitution{
 Autonomous Systems\\
 Daimler TSS\\
 Karlsruhe, Germany
}
\addinstitution{
 University Hospital Heidelberg\\
 Heidelberg, Germany
}
\addinstitution{
 Medical Image Computing \\
 German Cancer Research Center \\
 Heidelberg, Germany
}
\addinstitution{
 Institute for Artificial Intelligence in Medicine\\
 University Clinic Essen\\
 Essen, Germany
}

\runninghead{Seibold et al.}{Detailed X-Ray Annotations via CT Projection}
\def\eg{\emph{e.g}\bmvaOneDot}

\def\etal{\emph{et al}\bmvaOneDot}

\begin{document}

\maketitle

\begin{abstract}
    \noindent In clinical radiology reports, doctors capture important information about the patient's health status. They convey their observations from raw medical imaging data about the inner structures of a patient. As such, formulating reports requires medical experts to possess wide-ranging knowledge about anatomical regions with their normal, healthy appearance as well as the ability to recognize abnormalities. This explicit grasp on both the patient's anatomy and their appearance is missing in current medical image-processing systems as annotations are especially difficult to gather. This renders the models to be narrow experts \eg for identifying specific diseases.
    In this work, we recover this missing link by adding human anatomy into the mix and enable the association of content in medical reports to their occurrence in associated imagery (medical phrase grounding).
    To exploit anatomical structures in this scenario, we present a sophisticated automatic pipeline to gather and integrate human bodily structures from computed tomography datasets, which we incorporate in our \textsc{PAXRay}: A \underline{\textsc{P}}rojected dataset for the segmentation of \underline{\textsc{A}}natomical structures in \underline{\textsc{X-Ray}} data.  
    Our evaluation shows that methods that take advantage of anatomical information benefit heavily in visually grounding radiologists' findings, as our anatomical segmentations allow for up to absolute $50\%$ better grounding results on the OpenI dataset as compared to commonly used region proposals.
\end{abstract}
\setlength{\belowcaptionskip}{-6pt}
\addtolength{\abovecaptionskip}{-0pt}
\begin{figure}[t]
            \centering
            
            \begin{minipage}[c]{\linewidth}
              \centering
              \includegraphics[trim={0 0 0 0},clip,width=\linewidth]{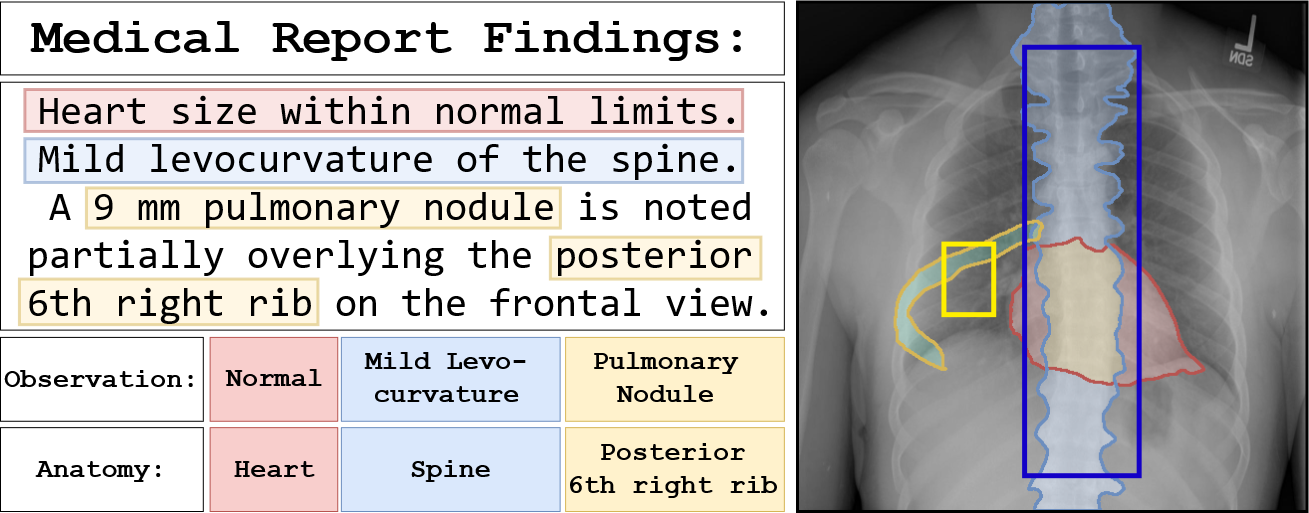}
              \caption{Overlap between the segmentation of anatomies and expert annotations on a sample of OpenI~\cite{openi} 
              indicating the necessity of anatomical understanding.
              Boxes are radiologists' annotation of findings. Masks show predictions for `6th right rib', `spine' and `heart'}
            \label{fig:label_eval2}
            \end{minipage}

        \label{fig:intro}
\end{figure}

\vspace{-0.25cm}
\section{Introduction}
With millions of images being produced every year, chest radiographs (CXR) are an essential part of daily clinical practice for initial diagnosis of pathologies such as rib fractures~\cite{awais2019diagnostic}, pneumothoraces~\cite{zarogoulidis2014pneumothorax} or pulmonary infections~\cite{borghesi2020covid}. 
For their interpretation, medical experts undergo extensive training to understand the present body structure and its consequent deviations for a radiologic image of a patient~\cite{brant2012fundamentals}. 
Subsequently, the radiologist summarizes the relevant visual information as a medical report for the further clinical workflow.  

In Fig.~\ref{fig:intro}, we display an example of a medical report. 
We display in the CXR on the right, that the radiolist's report follows anatomical structures to localize and describe anomalies similar to the prominent ABCDE-scheme~\cite{thim2012initial}. We argue the utilization between these correlations between anomalous findings and anatomical regions can be beneficial in the understanding of medical reports.
For example, the finding \textsc{\underline{pulmonary nodule} overlying the \underline{posterior sixth}} \textsc{\underline{rib}} can be localized using automatic anatomical segmentation. 

However, the arising challenge now becomes how to get hold of these segmentations? 
Dense annotations for natural~\cite{cordts2016cityscapes} and medical images~\cite{heller2019kits19} are challenging to collect.
For segmentations in X-rays, this issue is exacerbated due to the body absorbing radiation to a highly varying degree leading to anatomical structures in two-dimensional images being visibly overlayed with each other.
This leads to ambiguous, inextricable, visually blended patterns in X-rays that even with expert knowledge annotating fine-grained anatomy structures are unfeasible. This is also stated by Seibold~\etal~\cite{seibold2022reference} as the fine-grained mask annotation of a single CXR takes up to three hours. Due to this immense cost, most datasets stick to either a minimal mask labels~\cite{shiraishi2000development, jaeger2013automatic}, or strictly rely on image-level labels~\cite{wang2017chestx,bustos2020padchest,johnson2016mimic,openi,irvin2019chexpert}.

To bypass these issues, we find inspiration in three related facts: Firstly, computed tomography (CT) being aggregated multi-view 2D X-Rays~\cite{herman2009fundamentals}. 
Secondly, the immense advantages in identifying anatomy in CTs~\cite{chapman2016clinical}, i.e.~the esophagus can easily be tracked in a CT whereas it is harder in CXRs. Lastly, the consistent body structure of a patient throughout modalities. 
Building upon these observations, 
we contribute threefold:

We propose the use pipeline which makes use of proven segmentation methods in CTs to generate accurate anatomical annotation and subsequently transfers the 3D labels with the respective CT scan to 2D leading to simplified gathering of accurate CXR annotations.
    
Using this pipeline, we present the \emph{first} fine-grained anatomy dataset: \emph{PAXRay}. Based on high quality predictions in the CT space, we display 
92 individual labels of anatomical structures, which, when including super-classes, lead to a total of 166 labels in both lateral and frontal view.
 We make the dataset available for the community \href{https://constantinseibold.github.io/paxray/}{here}.
    
 Finally, we show that the usage of fine-grained anatomical structures can noticeably assist in matching medical observations and image regions. 
We, hereby, outperform commonly used region proposal methods by up to $50\%$ Hitrate for grounding methods.


\section{Related Work}
\noindent\textbf{Medical Image Understanding.}
The amount of CXR datasets \cite{openi,wang2017chestx,irvin2019chexpert,bustos2020padchest,johnson2019mimic} allowed for a massive development of deep learning approaches \cite{rajpurkar2017chexnet,wang2018tienet,seibold2020self,10.1007/978-3-030-86365-4_37,10.1007/978-3-030-87196-3_53,10.1007/978-3-030-87199-4_7,10.1007/978-3-030-87199-4_59,10.1007/978-3-030-87234-2_28}. These datasets are typically automatically annotated by a text classifier trained on a fixed set of diseases~\cite{irvin2019chexpert,smit2020chexbert,wang2017chestx}. Many works exist for the identification of diseases~\cite{wang2017chestx,rajpurkar2017chexnet,seibold2020self,bhalodia2021improving}, automated generation of reports~\cite{wang2018tienet,10.1007/978-3-030-87199-4_59} or visual question answering~\cite{sharma2021medfusenet}. While there have been methods which move away from fixed set training through multi-modal contrastive training to become more flexible~\cite{zhang2020contrastive, huang2021gloria, breaking, tiu2022expert}, deep learning algorithms in this area is widely regarded as a black box~\cite{burrell2016machine}. Several of these methods integrated interpretability through the use of class activation mappings~\cite{selvaraju2017grad,wang2017chestx,zhou2016learning} or attention~\cite{10.1007/978-3-030-87199-4_59} which, however, diverges from a doctor's anatomy-based approach~\cite{thim2012initial}. While some approaches emerged that utilize anatomical information~\cite{10.1007/978-3-030-87240-3_77,gordienko2018deep}, the level of detail is restricted to bounding boxes~\cite{wu2020automatic} or the heart and lung area as found in i.e. the JSRT dataset~\cite{shiraishi2000development}, thus narrowing down the potential field of application. Through the generation of our fine-grained PAX-Ray, the largest anatomy segmentation dataset at the time, we propose the usage of anatomical information in CXR to enable further interpretability of medical image analysis, and the diagnoses of physicians.

\noindent\textbf{Visual Phrase Grounding.}
Visual grounding seeks to encode informative content in natural language with visual features to localize visual content referenced in the text~\cite{deng2018visual,fukui2016multimodal,xiao2017weakly,datta2019align2ground,gonzalez2021panoptic,yang2019fast}. 
Most of such methods are two-stage methods~\cite{datta2019align2ground}. In the first stage, a region proposal method such as EdgeBoxes~\cite{zitnick2014edge}, Selective Search~\cite{uijlings2013selective} or trained detectors like Faster-RCNN~\cite{ren2015faster} generates potential regions of interest. In the second stage,  one tries to match queries to a fitting region based on their affinity~\cite{lee2018stacked,datta2019align2ground,gonzalez2021panoptic}. As the two-stage model performance directly relies on the usability of the proposal methods, they can be seen as their upper bound~\cite{yang2019fast}. In this work, we notice that proposal methods are suffering in the X-ray domain and propose to offset the shortcomings through the use of anatomical segmentations.
\vspace{-0.75cm}

\section{Automated Generation of Projected CXR Datasets}

\begin{figure*}[t]
    \centering
    \includegraphics[width=\linewidth, height=0.25\linewidth]{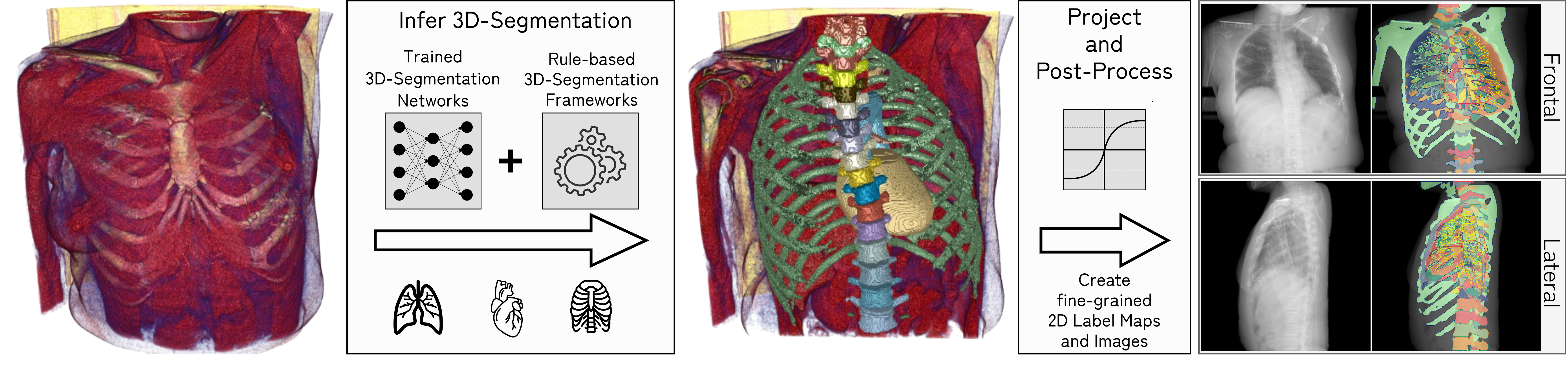}
    \caption{Dataset creation protocol. We apply established 3D segmentation methods to generate comprehensive annotations of a CT dataset. Afterwards, the CTs and their labels are projected to 2D using post-processing techniques to emulate X-ray characteristics.}
    \label{fig:dataset_creation}
\end{figure*}

Due to the immense difficulty of gathering precise pixel-wise annotations in the CXR domain, most datasets rely on either automatically parsed pathology labels or complete medical reports.
In contrast, as we extract information from much easier to annotate CTs, we propose a novel pipeline for generating annotations assisting the CXR domain and provide a densely labeled fine-grained dataset for anatomy segmentation containing both frontal and lateral views. Here, we leverage the consistency of anatomy between imaging domains to collect annotations from established models in the CT domain and then project these automatically generated annotations and images to 2D imitating the X-ray domain as shown in Fig.~\ref{fig:dataset_creation}.
\setlength{\belowcaptionskip}{-6pt}
\setlength{\abovecaptionskip}{-6pt}

\subsection{Automated Label Generation}\label{sec:label_mapping}
With the emergence of the UNet~\cite{unet,cciccek20163d,isensee2021nnu} the quality of 3D segmentation models for medical imaging has shown to be surprisingly reliable. For our annotation process, we utilize conventional segmentation methods ~\cite{lassen2011lung,lassen2012automatic} as well as the recent nnUNet~\cite{isensee2021nnu}.

We build the annotation scheme based on a label hierarchy with each label mask being denoted by $\mathcal{M}_{l}, l \in L$ with $L$ being the set of considered labels. We start with the generation of a body mask $\mathcal{M}_{body}$ to separate it from the CT-detector backplate by selecting the largest connected component after thresholding. We then consider the four super-categories of ~\textit{bones}, ~\textit{lung}, ~\textit{mediastinum} and ~\textit{sub-diaphragm}. Generally, we assume each fine-grained class as a subset of its parent class, i.e. the spine within the bone structure ($\mathcal{M}_{spine}  \subset {M}_{bone}$). We gather $\mathcal{M}_{bone}$ through a slicewise generalized histogram thresholding~\cite{barron2020generalization}. Within the bone structure, we segment the individual vertebrae~\cite{loffler2020vertebral,sekuboyina2021verse,liebl2021computed} and overall ribs~\cite{yang2021ribseg}. We expand the rib annotation of Yang~\etal\cite{yang2021ribseg} by discerning individual ribs as well as posterior and anterior parts based on their center and horizontal inflection.

For the lungs, we utilize the lung lobe segmentation model by Hofmanninger~\etal~\cite{hofmanninger2020automatic}. The merger of the individual lobes leads to the lung halves. We further gather the pulmonary vessels and total lungs through calculated thresholding and post-processing strategies~\cite{lassen2011lung}.

For the mediastinum, we considered the area between the lung halves. To segment this area we utilized Koitka~\etal's Body Composition Analysis (BCA) ~\cite{koitka2021fully} and split it into superior and inferior along the 4th T-spine following medical definitions~\cite{cook2015anatomy}. We extract annotations for the \textit{heart}, \textit{aorta}, \textit{airways}, and \textit{esophagus} using the SegThor dataset~\cite{lambert2020segthor}. 

As for the sub-diaphragm, we consider the area below the diaphragm. This area can be extracted from the soft tissue region segmented using the BCA which we split centrally into the left/right hemidiaphragm as no anatomical indicator exists.

To generate the label set $L$, we apply the combination of mentioned networks and rulesets on the publically available RibFrac~\cite{ribfrac2020} dataset which is fitting for such a projection process due to its focus on the thoracic area, the high axial resolution, and the scans being recorded without contrast agents similar to X-rays. We ignore volumes with contradicting segmentations and manually remove volumes with noticeable errors.

We project these labels to 2D using the $\max$-operation along the desired dimension and apply morphological post-processing steps based on the observed anatomy. We provide the full list of labels and the segmentation performance of the approaches in the supplementary.

\renewcommand{\arraystretch}{0.65}
\setlength{\tabcolsep}{1.5pt}
\setlength{\belowcaptionskip}{-6pt}
\setlength{\abovecaptionskip}{-2pt}

\begin{figure*}[t]
    \centering
        \begin{tabular}{ccccc}
        \toprule
         Projection & Lungs & Mediastinum & Bones & Sub-Diaphragm  \\
         \midrule
         \includegraphics[width=0.185\linewidth,height=0.185\linewidth]{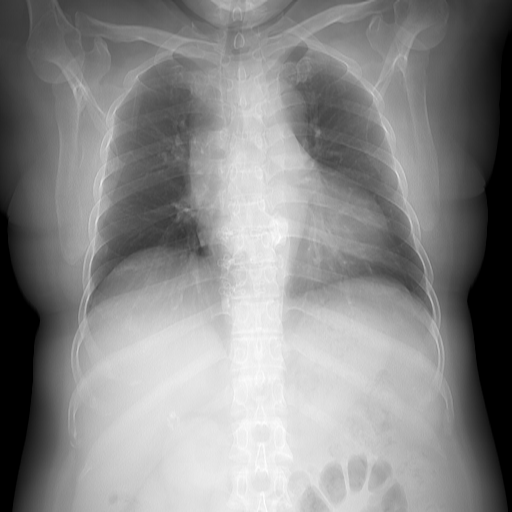}&
         \includegraphics[width=0.185\linewidth,height=0.185\linewidth]{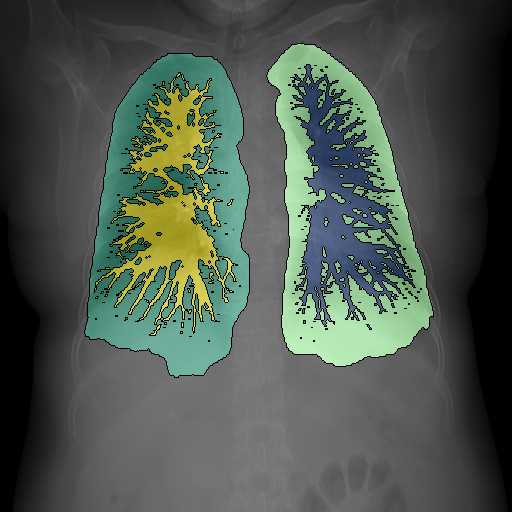}&
         \includegraphics[width=0.185\linewidth,height=0.185\linewidth]{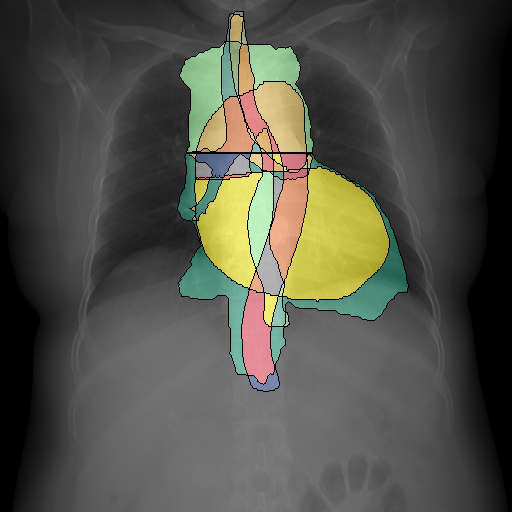}&
         \includegraphics[width=0.185\linewidth,height=0.185\linewidth]{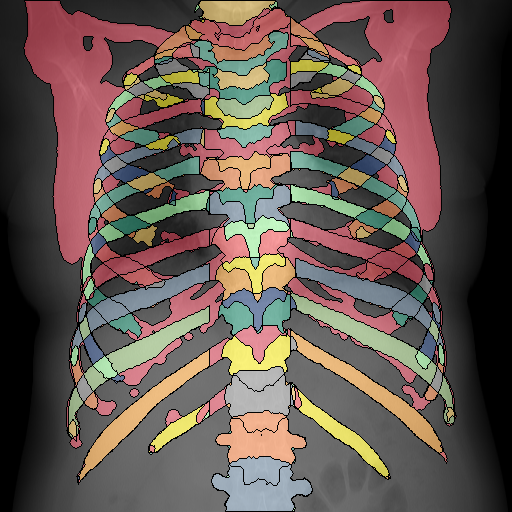}&
         \includegraphics[width=0.185\linewidth,height=0.185\linewidth]{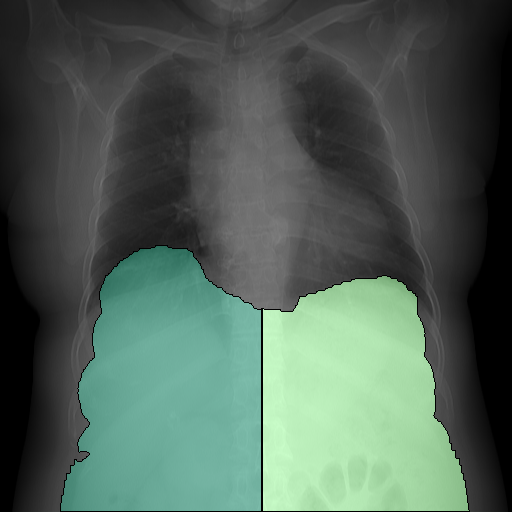}\\
         \includegraphics[width=0.185\linewidth,height=0.185\linewidth]{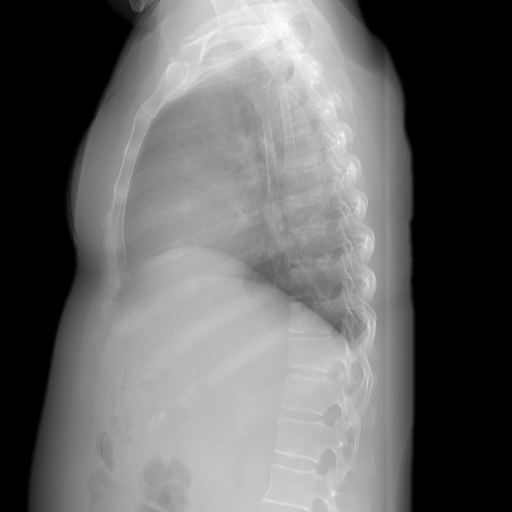}&
         \includegraphics[width=0.185\linewidth,height=0.185\linewidth]{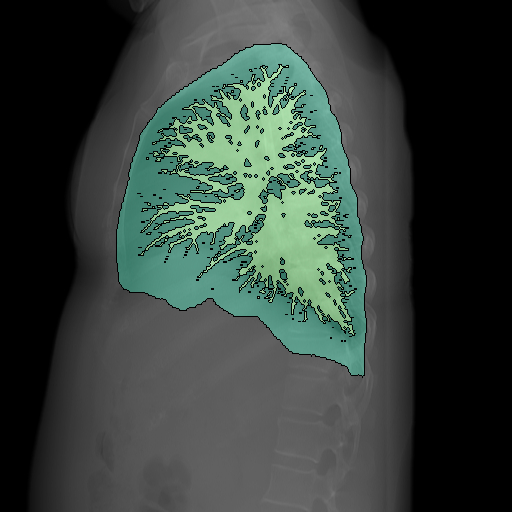}&
         \includegraphics[width=0.185\linewidth,height=0.185\linewidth]{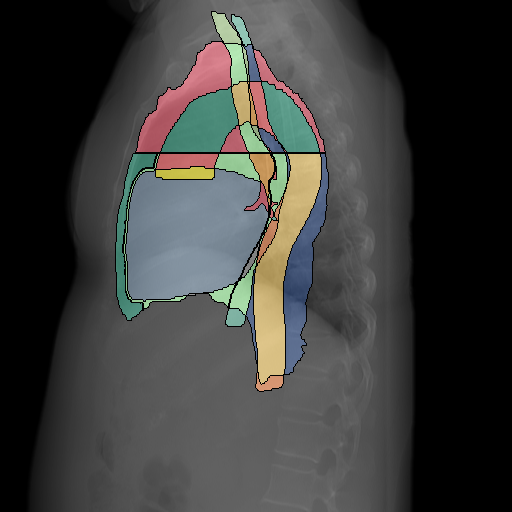}&
         \includegraphics[width=0.185\linewidth,height=0.185\linewidth]{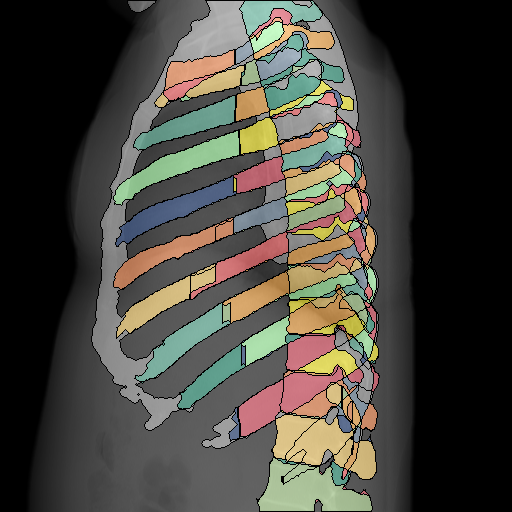}&
         \includegraphics[width=0.185\linewidth,height=0.185\linewidth]{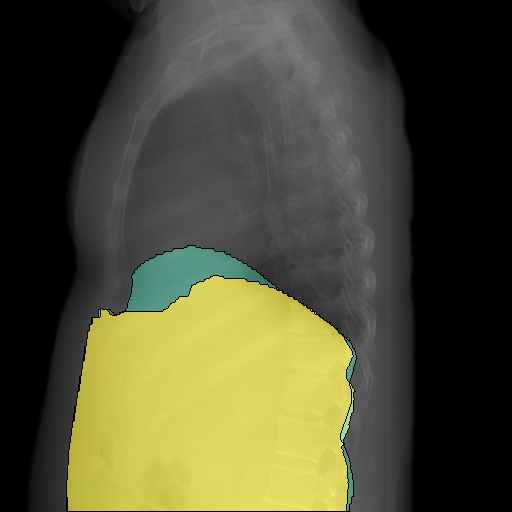}\\
         \bottomrule
    \end{tabular}
    \caption{Examples of overlapping annotations in both lateral and frontal view of PAX-ray}
    \label{fig:examples}
\end{figure*}

\subsection{CT to X-ray projection}
We project the CTs in similar to Kausch~\etal~\cite{10.1007/978-3-030-87202-1_34} and Matsubara~\etal ~\cite{matsubara2019generation}. 
Let $V$ be the volume of a CT scan and $M_{Body}, M_{Bone}$ the body and bone masks we gathered prior. 
We clip the $V$ to the common 12-bit range. We standardize the volume along the axis at which it is to be reduced, map it to the range of $[0,1]$ via a sigmoid function $\sigma$ and sharpen:
\begin{equation}
     V_{Body}' = M_{Body} \cdot \sigma\left(\frac{V-mean(V)}{std(V)}\right).
\end{equation}
We repeat this for the bone region to get $V'_{Bone}$. Afterwards, the $V'$s are summed, min-max-feature scaled, and rescaled to the desired range. We average along the desired dimension to get the image.
We show exemplary image-label-pairs in Fig.~\ref{fig:examples} and the supplemental.

\section{Anatomy-guided Phrase Grounding of Medical Reports}
\begin{figure*}[t]
    \centering
    \includegraphics[width=\linewidth]{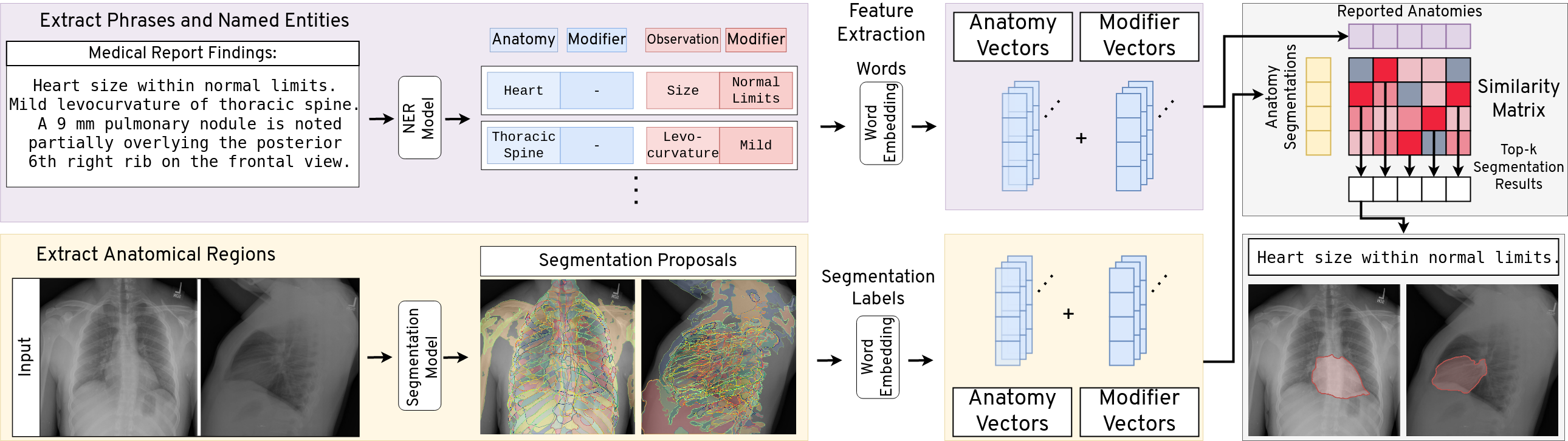}
    \caption{Anatomy Grounding Baseline. Using NER divide phrases into \textit{Anatomy}, \textit{Anatomy-Modifier}, \textit{Observation}, \textit{Observation-Modifier}. For the images, we generate proposals using our anatomy segmentation model. We extract word-wise embeddings for the phrase/anatomy labels, aggregate them and retrieve the most similar region for each phrase}
    \label{fig:method}
\end{figure*}

In medical reports, oftentimes medical observations are paired with anatomical regions to refer to their respective positions. Starting from the assumption of co-occurrence between diseases and anatomical regions within the text, we build a straightforward baseline to indicate the usability of anatomy guidance for the grounding of observations as seen in Fig.~\ref{fig:method}.

For each image-report pair $(\mathcal{I}_i,\mathcal{R}_i) \in \{(\mathcal{I}_1,\mathcal{R}_1), (\mathcal{I}_2,\mathcal{R}_2), \dots, (\mathcal{I}_N,\mathcal{R}_N)\}$ in a dataset consisting of $N$ pairs, we consider the \emph{finding}-section of the report containing the description of visual observations in the image. As shown in the top branch of Fig.~\ref{fig:method}, we process the report $\mathcal{R}_i$ sentence-wise to split it into medically relevant phrases $\mathcal{P}_{ij} \in \mathcal{P}_{i}, 0\leq j \leq  \lvert \mathcal{P}_i \lvert$ that are classified as problem or treatment by the named-entity-recognition (NER) model and discard the rest \cite{stanza,johnson2016mimic,uzuner20112010}.
Subsequently, we filter the words $w \in \mathcal{P}_{ij}$ using the NER-model $\mathcal{C}$ to classify $w$ into \textit{Anatomy} $A$~(\eg heart, \dots), \textit{Anatomy-modifier} $AM$~(\eg posterior,\dots), \textit{Observation} $O$~(\eg pneumothorax, \dots) or \textit{Observation-modifier} $OM$~(\eg above, \dots)~\cite{stanza,johnson2016mimic}, group them and omit $w$ if it doesn't belong to any of these categories.
Thus, we get a filtered phrase $\mathcal{P}_{ij}^*$ which contains groups of the words $W^{x}_{ij}$ of each category $x \in \{A, AM, O, OM\}$:
\begin{equation}
\begin{aligned}
    \mathcal{P}_{ij}^* &= \{W_{ij}^A, W_{ij}^{AM}, W_{ij}^{O}, W_{ij}^{OM}\} \\ W_{ij}^x &= \{w \mid \mathcal{C}(w) = x, w \in \mathcal{P}_{ij}\}, 
\end{aligned}
\end{equation}
We utilize a pre-trained word-embedding model $\mathcal{E}$ to extract $d$-dimensional embeddings for all words in the filtered phrase $\mathcal{P}_{ij}^*$ occurring as anatomy and anatomy modifier: 
\begin{equation}
\begin{aligned}
    F^A_{ij} &= \{\mathcal{E}(w)\mid w \in W^A_{ij}\} \\
    F^{AM}_{ij} &= \{\mathcal{E}(w)\mid w \in W^{AM}_{ij}\}
\end{aligned}
\end{equation} For phrases occurring without an anatomy or anatomy modifier, we set $F^{x}_{ij} = 0^d$ with $x\in \{A, AM\}$. As  multiple words for a phrase can occur as anatomy or anatomy modifier we consider the category representation as mean of all word embeddings belonging to that specific category. 
\setlength{\belowcaptionskip}{-6pt}
\setlength{\abovecaptionskip}{-6pt}

The final phrase embedding is then the sum of both category embeddings. 
\begin{equation}
    F_{ij} = \text{mean}(F^{A}_{ij}) + \text{mean}(F^{AM}_{ij})
\end{equation}

In the bottom branch of \cref{fig:method}, we extract anatomical regions using our segmentation network. Doing so we get 166 binary predictions with their associated class label text $l \in L$ for each view, which we threshold to get mask regions. We refine these segmentation masks through similar anatomical constraints of their parent classes and post-processing steps as in Section~\ref{sec:label_mapping}.
For all segmentation masks we split their label text $l$ into anatomy and its modifier and we extract features $T_{l}$ in a similar manner as above: 
\begin{equation}
\begin{aligned}
     T^A_{l} &= \{\mathcal{E}(w)\mid \mathcal{C}(w) = A, w \in l \}\\
     T^{AM}_{l} &= \{\mathcal{E}(w)\mid \mathcal{C}(w) = AM, w \in l \}\\
     T_{l} &= mean(T^{A}_{l}) + mean(T^{AM}_{l})
\end{aligned}
\end{equation}
Utilizing these feature vectors, we compute the cosine-similarity matrix $S^i \in [-1,1]^{|\mathcal{P}_i| \times 166}$ between both image regions and phrases individually for the lateral- and frontal view with each entry being defined by $S^i_{jl} = cos(F_{ij},T_{l})$.
Then, for a given phrase query we return the segmentation proposal based on the top-$k$ similarities. 
For phrases without anatomy we simply return the whole image. 

\subsection{Implementation Details}
\noindent\textbf{Anatomy Segmentation:}
To show the usability of our fine-grained multi-label dataset, we train segmentation models with differently trained backbone networks. We chose the in the medical domain commonly utilized UNet\cite{unet} and the SFPN\cite{kirillov2019panoptic} with a ResNet-50\cite{he2016deep} backbone. As the labels can overlap we train with binary cross-entropy and employ an additional binary dice loss. We used random resize-and-cropping of range $[0.8,1.2]$ as augmentation with an image size of 512 and optimize using AdamW\cite{loshchilov2017decoupled} with a learning rate of 0.001 for 110 epochs decaying by a factor of 10 at $\{60,90,100\}$ epochs.

\noindent\textbf{Phrase Grounding:}
We process our reports using Stanza~\cite{zhang2021biomedical} to infer observations/treatments using the i2b2-2010 corpus~\cite{uzuner20112010} as well as anatomies and observations through the Radiology corpus~\cite{zhang2021biomedical,johnson2016mimic}.
We utilize ChexBert~\cite{smit2020chexbert} to extract an additional 
\textit{is-anomaly} 
token for each phrase. 
To extract word and phrase features we utilize BioWordVec~\cite{zhang2019biowordvec} and
BioSentVec~\cite{chen2019biosentvec}.
As we evaluate grounding in this task via bounding box comparison, for each segmentation result we extract a corresponding bounding box.

\renewcommand{\arraystretch}{0.65}
\setlength{\tabcolsep}{1.5pt}
\setlength{\belowcaptionskip}{-14pt}
\setlength{\abovecaptionskip}{0pt}
\begin{figure*}[t]
    \centering
        \begin{tabular}{cccccc}
        \toprule
         &Input & Lungs & Mediastinum & Bones & Sub-Diaphragm  \\
         \midrule
         \rotatebox{90}{$\phantom{00}$ Prediction}&\includegraphics[width=0.18\linewidth,height=0.18\linewidth]{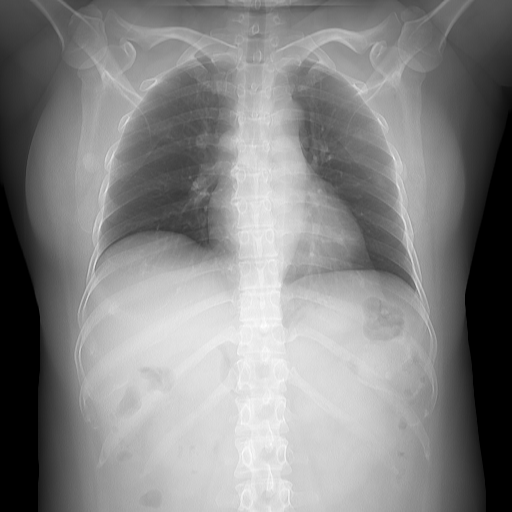}&
         \includegraphics[width=0.18\linewidth,height=0.18\linewidth]{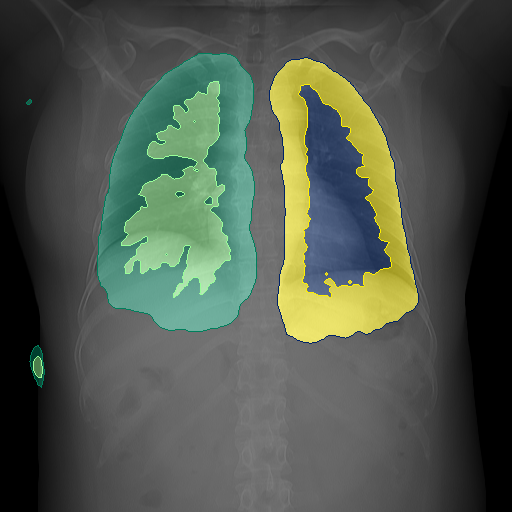}&
         \includegraphics[width=0.18\linewidth,height=0.18\linewidth]{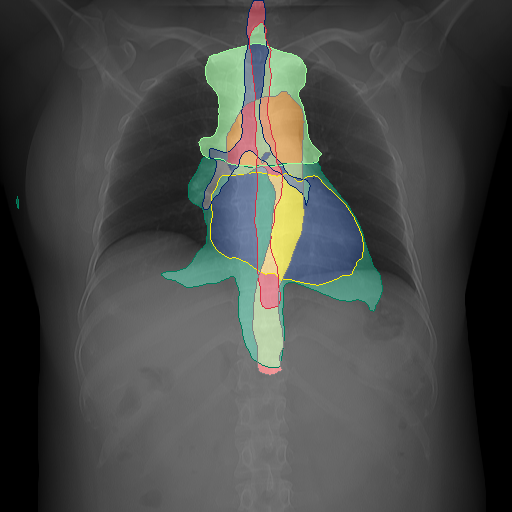}&
         \includegraphics[width=0.18\linewidth,height=0.18\linewidth]{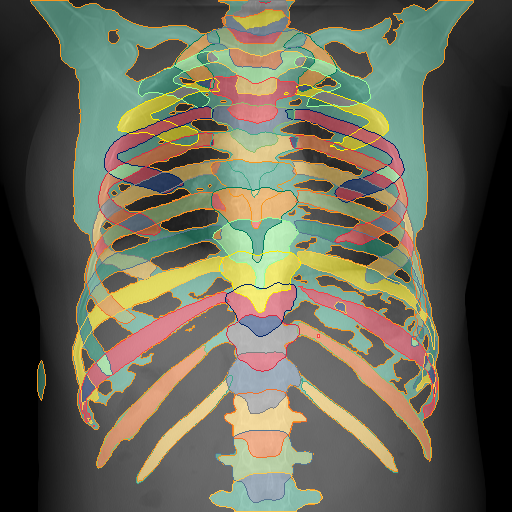}&
         \includegraphics[width=0.18\linewidth,height=0.18\linewidth]{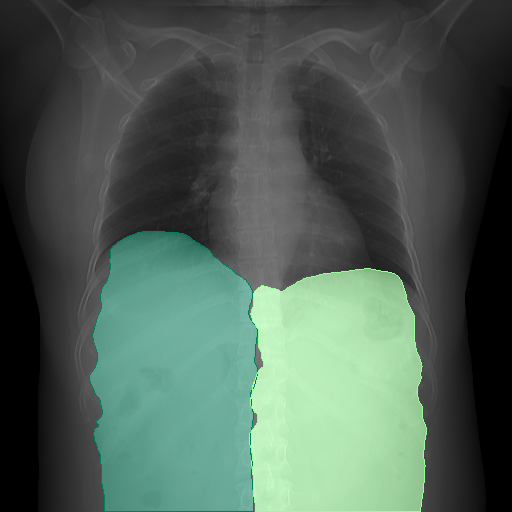}\\
         \rotatebox{90}{$\phantom{000}$ Target}&\includegraphics[width=0.18\linewidth,height=0.18\linewidth]{images/pred_panda/RibFrac484frontal.png}&
         \includegraphics[width=0.18\linewidth,height=0.18\linewidth]{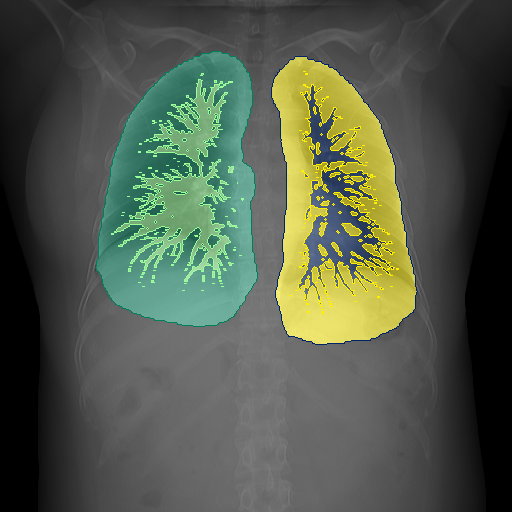}&
         \includegraphics[width=0.18\linewidth,height=0.18\linewidth]{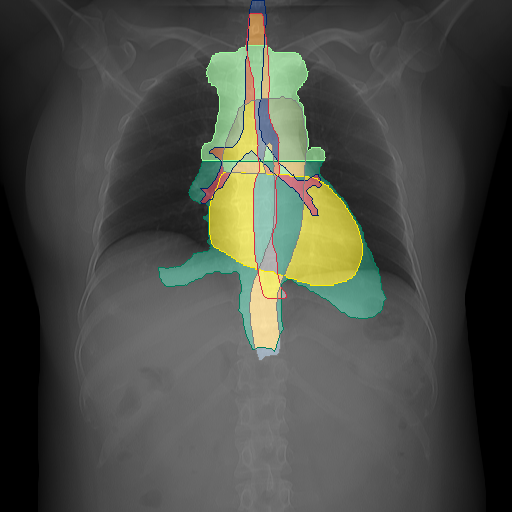}&
         \includegraphics[width=0.18\linewidth,height=0.18\linewidth]{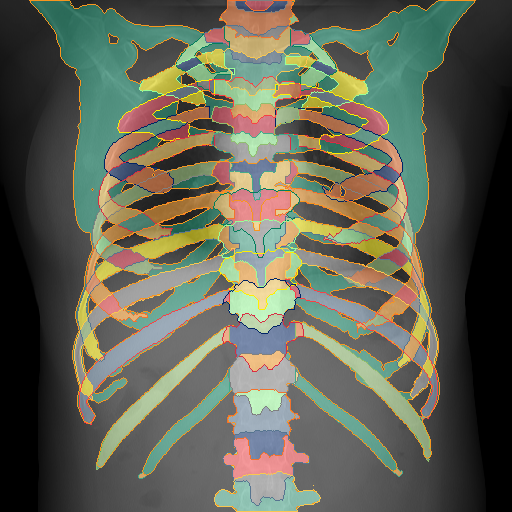}&
         \includegraphics[width=0.18\linewidth,height=0.18\linewidth]{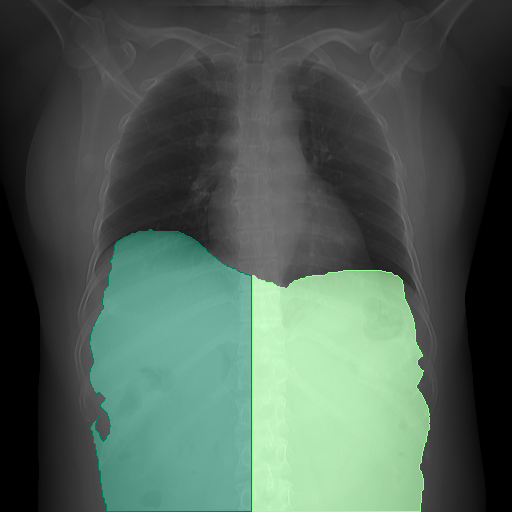}\\
         \bottomrule
    \end{tabular}
    \caption{Qualitative results of a UNet on the test set of the \emph{PAXRay} dataset}
    \label{fig:pred_panda}
\end{figure*}
\renewcommand{\arraystretch}{0.4}
\setlength{\tabcolsep}{1.25pt}
\setlength{\belowcaptionskip}{-6pt}
\setlength{\abovecaptionskip}{0pt}
\begin{figure*}[t]
    \centering
        \begin{tabular}{ccccc}
        \toprule
         Input & Lungs & Mediastinum & Bones & Sub-Diaphragm  \\
         \midrule
         \includegraphics[width=0.18\linewidth,height=0.18\linewidth]{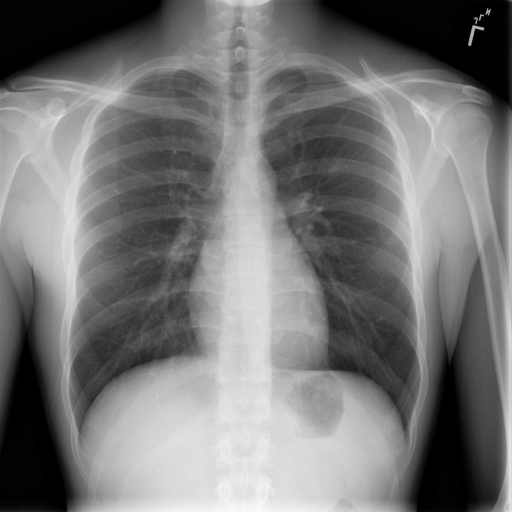}&
         \includegraphics[width=0.18\linewidth,height=0.18\linewidth]{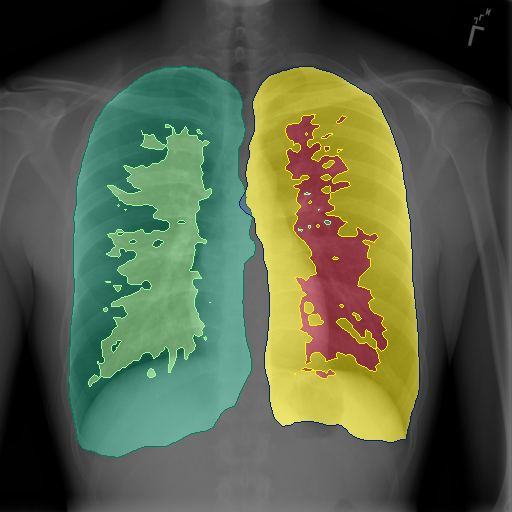}&
         \includegraphics[width=0.18\linewidth,height=0.18\linewidth]{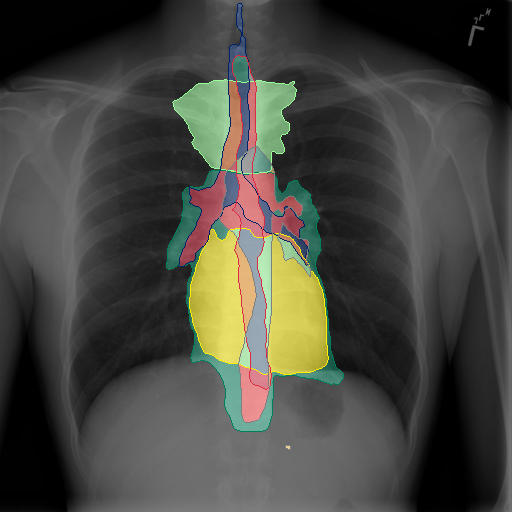}&
         \includegraphics[width=0.18\linewidth,height=0.18\linewidth]{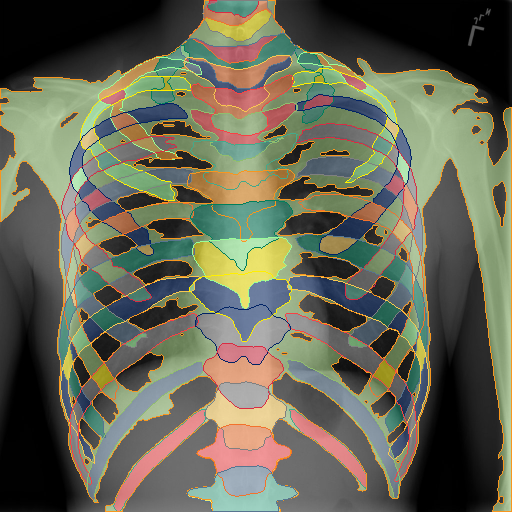}&
         \includegraphics[width=0.18\linewidth,height=0.18\linewidth]{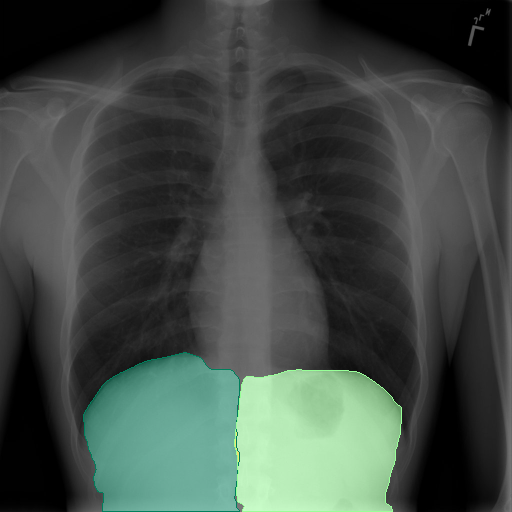}\\
         \includegraphics[width=0.18\linewidth,height=0.18\linewidth]{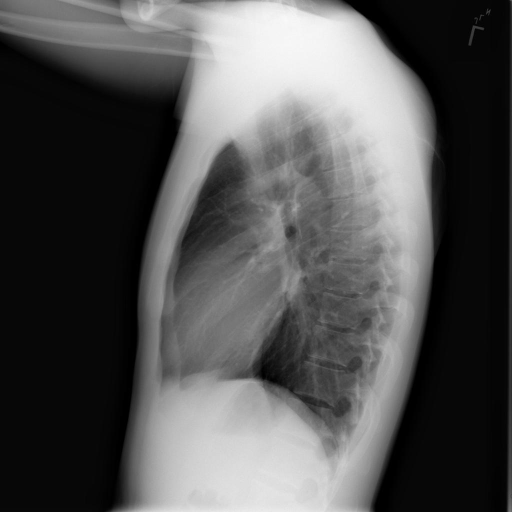}&
         \includegraphics[width=0.18\linewidth,height=0.18\linewidth]{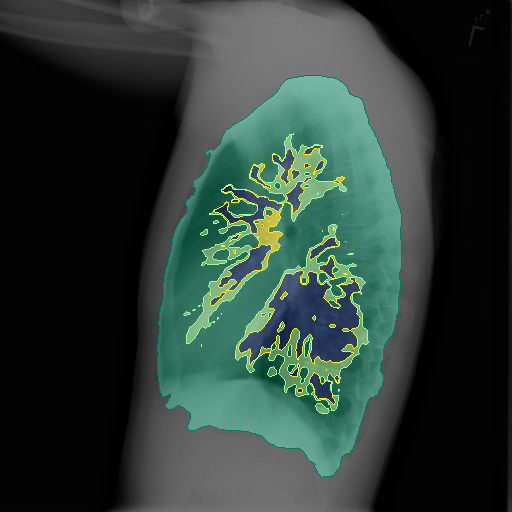}&
         \includegraphics[width=0.18\linewidth,height=0.18\linewidth]{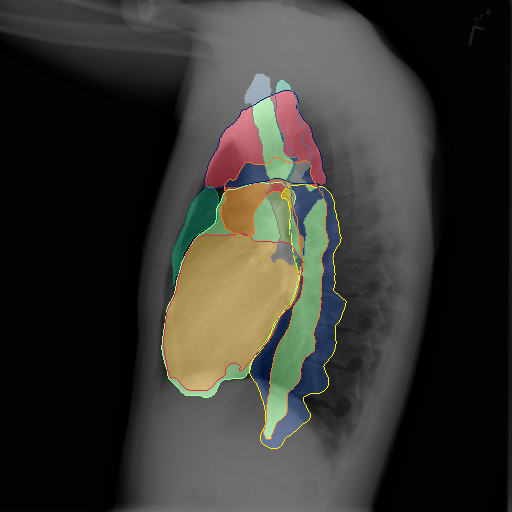}&
         \includegraphics[width=0.18\linewidth,height=0.18\linewidth]{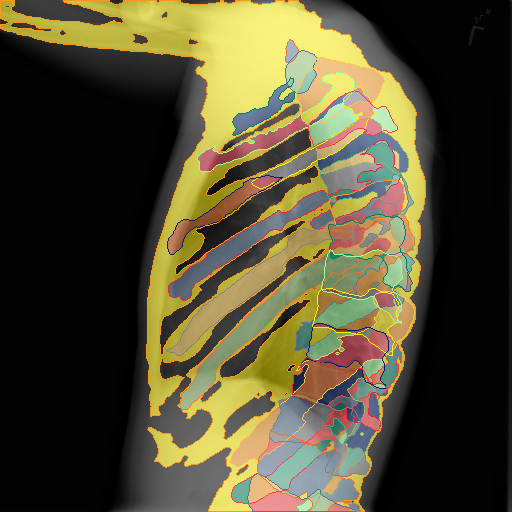}&
         \includegraphics[width=0.18\linewidth,height=0.18\linewidth]{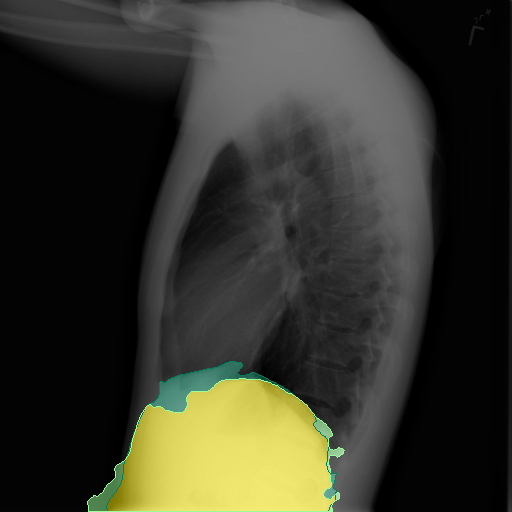}\\
         \bottomrule
    \end{tabular}
    \caption{Qualitative results of a UNet trained on our PAX-ray dataset for a patient in OpenI}
    \label{fig:segmentation_results}
\end{figure*}

\section{Experiments}
\subsection{Anatomy Segmentation}\label{sec:eval_segmentation}
\noindent{\textbf{Experimental Setting:}} We evaluate the segmentation quality quantitatively on the PAXRay dataset using the typically used mean Intersection over Union (IoU). We maintain the train/ val/test splits of the RibFrac dataset~\cite{ribfrac2020} as such we are left with 598/74/180 images. We validate our models every 10th epoch and test the model which performed best on validation. 

\renewcommand{\arraystretch}{1.125}
\setlength{\tabcolsep}{1.1pt}
\begin{table*}[b]
    \centering
\begin{tabular}{ cc c cc c cccc c cc c c ccccc}
\toprule
 \multirow{2}{*}{} &  \multirow{2}{*}{Init.} & \phantom{a} &
 \multicolumn{2}{c}{Lung} & \phantom{a} & \multicolumn{4}{c}{Mediastinum} & \phantom{a} & \multicolumn{2}{c}{Bones}& \phantom{a} & \multirow{2}{*}{Sub-Dia.} & \multirow{2}{*}{Mean}\\

 \cmidrule{4-5} \cmidrule{7-10} \cmidrule{12-13}
 
      &          & \phantom{a} & Lobes & Vessels & \phantom{a} &Regions & Heart & Aorta & Airw. & \phantom{a} & Spine & Ribs & \\
\midrule
 
\multirow{2}{*}{\rotatebox{90}{\makecell{SFPN}} } & (Random) && 82.3
& 49.5
&& 68.6
& 81.8
& 67.8
& 55.6&
&84.8
& 69.4&
& 93.9
& 37.8\\


&  (VBData) 
&& \text{86.3}
& \textbf{52.1}
&& 74.6
& \text{88.9}
& \text{79.0}
& 70.0 &
& \text{90.5}
& \text{78.8}&
& \text{96.2}
& \text{51.9}\\

\midrule

\multirow{2}{*}{\rotatebox{90}{\makecell{UNet}} }&  (Random) & 
& 85.0 & 49.8 && 74.8 
& 87.7 & 77.9 & 68.8 && 90.0 & 81.5 && 95.6 & 54.5 \\
 &  (VBData) && \textbf{86.9}
 & 50.8 &
 & \textbf{77.3}
 & \textbf{89.9} & \textbf{80.8} & \textbf{73.2} && \textbf{92.5} & \textbf{84.9} & & \textbf{96.7} & \textbf{60.6}\\
  \bottomrule
\end{tabular}
    \caption{Segmentation performance in IoU on the test split of our proposed \emph{PAXRay} dataset}
    \label{tab:segmentation_results}
\end{table*}

\noindent\textbf{Results:}  We see quantitative results in Table~\ref{tab:segmentation_results}. We show the performance on selected super-classes and the mean over all 166 classes due to the immense number of classes. We show the complete performance over all classes in the supplementary material.

We see that in the observed setting we profit from pre-trained networks with a gain from up to $\sim14\,\%$ mIoU. We see a difference in performance based on the architecture as the UNet outperforms the SFPN  by roughly $9\,\%$. While for classes like the heart, lobes, or aorta these architectures perform similarly there are noticeable differences for individual mediastinal regions, airways, and ribs.

While we are able to segment several classes with up to $\sim90\,\%$ mIoU on classes like the spine or heart, the correct segmentation of lung vessels is especially difficult with an IoU of $52\,\%$. Furthermore, the largest difference in segmentation quality between the two networks lies in the rib cage segmentation where the UNet has a gain of $6.1\%$. We observe qualitative examples in Fig.~\ref{fig:pred_panda}. The vessel tree and tracheal ends towards the bronchi pose as difficult, whereas lobe-, intermediastinal-, and bone-related classes appear as expected. 

To show the applicability of our proposed dataset for the anatomy segmentation in real CXR, we display qualitative results on the OpenI dataset in Fig.~\ref{fig:segmentation_results}. 
While similar errors to the projected x-rays can be noticed i.e. for the lateral rib or diaphragm segmentations, the results show to be quite promising albeit no domain adaptation method~\cite{toldo2020unsupervised} was applied. 

\renewcommand{\arraystretch}{1.0}
\setlength{\tabcolsep}{1pt}
\setlength{\belowcaptionskip}{-10pt}
\setlength{\abovecaptionskip}{-0pt}
\begin{table}[t]
    \centering
    
    \begin{tabular}{ccc}
         \begin{tabular}{ c c c c c}
\toprule
& Method(N=200) & $\text{HR}_{25}$ & $\text{HR}_{50}$ & $\text{HR} _{75}$ \\
\midrule
\multirow{5}{*}{\rotatebox{90}{\makecell{Frontal}}} &  Whole Image  & 16.5 
& 5.8
& 0.4  \\
& Selec. Search  \cite{uijlings2013selective} &  \text{72.8} & 16.5 & \text{7.7}  \\
& EdgeBoxes \cite{zitnick2014edge}  &  18.9 & 4.8 & 0.9  \\
& RPN \cite{ren2015faster} &  53.8 & \text{18.9} & 1.4  \\
& Anatomy Segm. &  \textbf{93.2} & \textbf{66.9} & \textbf{20.8}  \\
\bottomrule
\end{tabular}
         & $\phantom{0}$ &
         \begin{tabular}{ c c c c c}
\toprule
& Method(N=200) & $\text{HR}_{25}$ & $\text{HR}_{50}$ & $\text{HR} _{75}$ \\
\midrule
\multirow{5}{*}{\rotatebox{90}{\makecell{Lateral}}} & Whole Image & 23.1
& 8.4 
& 1.0\\

& Selec. Search \cite{uijlings2013selective}  &  \text{80.7} & \text{47.7} & \text{19.2}  \\
& EdgeBoxes \cite{zitnick2014edge}  & 35.7& 11.9 & 1.8  \\
& RPN \cite{ren2015faster} &  68.8 & 24.7 & 0.9  \\
& Anatomy Segm. &  \textbf{88.0} & \textbf{62.3} & \textbf{20.1}  \\
\bottomrule
\end{tabular}
    \end{tabular}
\caption{Hit rates of region proposals for different IoU thresholds (denoted by subscript).}
    \label{tab:hitrate}
\end{table}

\subsection{Medical Phrase Grounding}\label{sec:eval_grounding}
\noindent\textbf{Experimental setting: }For our evaluation of medical phrase grounding, we use the OpenI dataset~\cite{openi} which consists of medical reports paired with frontal and lateral chest X-rays. We tasked two radiologists to highlight phrases within 100 medical reports in the OpenI dataset resulting in 178 frontal and 146 lateral bounding box annotations. 

We evaluate the usability of anatomy segmentations for medical phrase grounding in two parts. First, we investigate the upper bound achievable by computing the average hit rate (HR) at different IoU thresholds~\cite{yang2019fast}. A hit is considered as a candidate region overlapping with the ground truth annotation with an IoU above the set threshold. We compare our anatomy segmentations with common region proposal methods utilized by phrase grounding algorithms~\cite{datta2019align2ground,moradi2018bimodal,fukui2016multimodal} in natural images such as EdgeBoxes~\cite{zitnick2014edge}, Selective Search~\cite{uijlings2013selective} and Region Proposal networks~\cite{ren2015faster}. We extract the top 200 scoring boxes for each labeled image following most phrase grounding methods~\cite{datta2019align2ground,fukui2016multimodal,yang2019fast}.

\renewcommand{\arraystretch}{1.0}
\setlength{\tabcolsep}{1pt}
\setlength{\belowcaptionskip}{-5pt}
\setlength{\abovecaptionskip}{0pt}

\begin{table*}[b!]
    \centering
    
\begin{tabular}{ c c cc ccccc}
\toprule
& \multirow{2}{*}{Method}  & \multirow{2}{*}{\makecell{Box\\Proposals}} & \multirow{2}{*}{\makecell{Text\\Features}} & \multirow{2}{*}{\makecell{$\text{Top-1}_{25}$ }} & 
\multirow{2}{*}{\makecell{$\text{Top-1}_{50}$ }} &
\multirow{2}{*}{\makecell{$\text{Top-1}_{75}$ }} &
\multirow{2}{*}{\makecell{$\text{Top-5}_{50}$ }} &
\multirow{2}{*}{\makecell{$\text{Top-10}_{50}$ }} \\
&&& &&  & \\
\midrule
\multirow{6}{*}{\rotatebox{90}{\makecell{Frontal}}} & Whole Image  & None & None & 18.5&7.1&0.5&7.1& 7.1\\
& \dashuline{Oracle} & Sel. Search & None &  \dashuline{72.8}  &  \dashuline{16.5}  & \dashuline{7.7} & \dashuline{16.5} & \dashuline{16.5}  \\
& PhraseDist &  Anat. Seg. & BioSent &  36.5  &  17.9  & 2.9& 23.3 & 27.5  \\
& Anat.Dist &  Anat. Seg. & BioWord &  34.7  &  13.1  & 0.5  & 26.3 & \textbf{28.1} \\
& ModAnat. &  Anat. Seg. & BioWord &  \textbf{38.9}  &  \textbf{21.5}  & \textbf{4.7} & \textbf{27.5} & \textbf{28.1}  \\

\midrule
 \multirow{6}{*}{\rotatebox{90}{\makecell{Lateral}}} & Whole Image & None & None&  23.1 & 8.4 & 1.0 & 8.4 & 8.4\\
& \dashuline{Oracle} &  Sel. Search & None &  \dashuline{80.7}  &  \dashuline{47.7}  & \dashuline{19.2} & \dashuline{47.7} & \dashuline{47.7}  \\
& PhraseDist &  Anat.Seg. & BioSent &  47.3  &  22.1  & 4.2 & 26.3 & 30.5  \\
& Anat.Dist. &  Anat.Seg. & BioWord & 45.2  &  17.8  & 2.1  & 30.5 & 31.5 \\
& ModAnat. &  Anat. Seg. & BioWord &  \textbf{49.4}  &  \textbf{26.3}  & \textbf{8.4}  & \textbf{32.6} & \textbf{32.6} \\
 \bottomrule
 
\end{tabular}
\caption{Medical phrase grounding performance on OpenI showing Top-k region retrieval performance at different IoU thresholds (denoted by the subscript).}
    \label{tab:grounding}
\end{table*}

Afterwards, we show the performance of our proposed baseline in terms of Top-1/5/10 region retrieval at IoU thresholds of 25/50/75\,\% and compare it to using the entire phrase for the comparison with our label as well as just the anatomy in itself. We also display the \emph{oracle}'s performance utilizing selective search to put the value of the proposed anatomy-based segmentations into perspective as it poses as the upper bound of weakly supervised methods, i.e. if the proposal method is unable to provide good initial hints the grounding method itself cannot match phrases with their image region.  


\noindent\textbf{Hit Rate Analysis:}  We show hit rate (HR) results in Table~\ref{tab:hitrate}. We see that for the traditional approaches in both the frontal and the lateral view the selective search algorithm
provided the best proposals, however, we observe an extreme loss in quality when increasing the IoU 
threshold, i.e. in the frontal view the 
hit rate drops
by nearly $56\,\%$. These $16.5\,\%$ stand in  comparison to 
the Flicker30K  
dataset where the hit rate of selective search at a $50\,\%$ IoU threshold for 200 boxes was reported as $85.68\,\%$ ~\cite{yang2019fast}. 
In contrast, without being trained in the segmentation of observations but rather anatomies, we achieve improvements across all categories with \ie a $50\,\%$ improvement in HR for the frontal view at an IoU of $50\,\%$. This indicates that anatomy guidance can be a better starting point for the localization of observations as the HR relates to the oracle's performance.

\begin{figure*}[t]
    \centering
         \includegraphics[width=\linewidth]{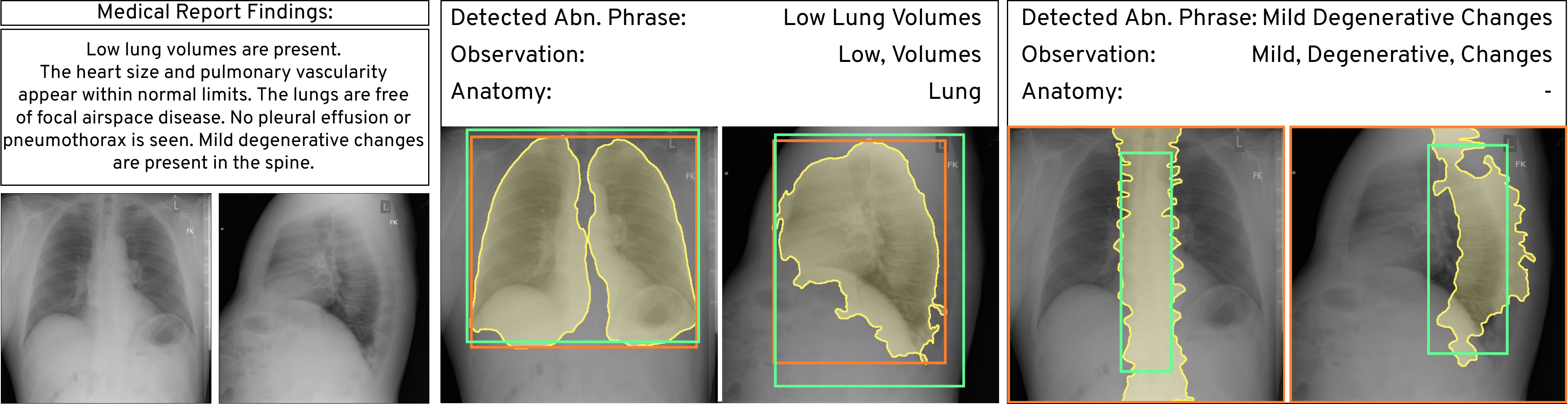}
    \caption{We show \textcolor{green}{ground truth}, \textcolor{orange}{retrievals} and \textcolor{yellow}{expected retrieval}. If anatomy phrases are identified a result is provided. Otherwise, the segmentation, albeit accurate, is not retrieved.  }
    \label{fig:qualitative_results}
\end{figure*}

\noindent\textbf{Grounding Results:} We show our quantitative results for medical phrase grounding in Table~\ref{tab:grounding}. We see that both the direct sentence comparison as well as  our proposed method surpass the oracle's performance based on proposals by selective search for the frontal view on the commonly used IoU threshold of $50\,\%$. Utilizing both anatomy and their modifiers improves noticeably over using complete sentence embeddings. 
We show qualitative results in Figure~\ref{fig:qualitative_results}. We highlight anatomy and medical phrases. We see that despite not directly referring to disease, anatomical regions can be utilized to retrieve medical findings.

\setlength{\belowcaptionskip}{-6pt}
\setlength{\abovecaptionskip}{-6pt}
\renewcommand{\arraystretch}{0.65}
\setlength{\tabcolsep}{1.5pt}

\vspace{-0.5cm}
\section{Discussion and Conclusion} 
In this paper, we propose a method for the automatic generation of anatomy labels for chest X-rays through the projection of CT data and their respective annotations via established segmentation methods to enable more complex downstream tasks such as medical phrase grounding. 
As the required time for fine-grained mask annotations in medical images is massive, our scheme can be considered to be an immense time save for the generation of CXR annotations and could be extended to any annotation type, be it anatomy or pathology. 

We introduce the PAXRay dataset which consists of projected CXR paired with a large amount of fine-grained anatomical structures. 
The information richness of dense pixel-wise annotations and the shared anatomical context between X-Rays allows us to train fine-grained anatomy segmentation models. 
Furthermore, with our proposed method the PAXRay dataset can be extended arbitrarily by utilizing additional CT datasets. 
We show in our experiments that the resulting models can segment anatomical regions on not only projected but also real CXR images, thus, enabling  precise anatomy localization to build reliable region proposals for CXR analysis.
This allows us to outperform prior oracle-like performance with a simple baseline method. 
We anticipate that our work allows the community to develop improved methods for generating more interpretable computer-assisted diagnosis tools.


\section{Acknowledgements}
The present contribution is supported by the Helmholtz Association under the
joint research school “HIDSS4Health – Helmholtz Information and Data Science
School for Health”.

\bibliography{egbib}
\end{document}